\documentclass[11pt,a4paper]{article}
\usepackage[hyperref]{acl2020}
\usepackage{times}
\usepackage{latexsym}

\usepackage{microtype}

\usepackage{times}
\usepackage{latexsym}
\usepackage{color}
\usepackage{tabularx}
\usepackage{multirow}
\usepackage{caption}
\usepackage{subcaption}
\usepackage{tikz}
\usepackage{tikz-qtree}
\usepackage{url}
\usepackage{color, colortbl}
\usepackage{amsmath}
\usepackage{enumitem}
\usepackage{caption}
\newcommand{\heading}[1]{\multicolumn{1}{c}{#1}}
\usepackage{pgfplots}
\pgfplotsset{compat=1.14}

\aclfinalcopy 

\title{Is this Dialogue Coherent? Learning from Dialogue Acts and Entities}

\author{Alessandra~Cervone and Giuseppe Riccardi \\
  Signals and Interactive Systems Lab, University of Trento, Italy \\
  \texttt{{\{alessandra.cervone, giuseppe.riccardi\}}@unitn.it} \\
  }

\date{}

\begin{document}
\maketitle
\begin{abstract}
In this work, we investigate the human perception of coherence in open-domain dialogues. In particular, we address the problem of annotating and modeling the coherence of next-turn candidates while considering the entire history of the dialogue.  
First, we create the Switchboard Coherence (SWBD-Coh) corpus, a dataset of human-human spoken dialogues annotated with turn coherence ratings, where next-turn candidate utterances ratings are provided considering the full dialogue context.
Our statistical analysis of the corpus indicates how turn coherence perception is affected by patterns of distribution of entities previously introduced and the Dialogue Acts used.
Second, we experiment with different architectures to model entities, Dialogue Acts and their combination and evaluate their performance in predicting human coherence ratings on SWBD-Coh. 
We find that models combining both DA and entity information yield the best performances both for response selection and turn coherence rating.
\end{abstract}

\section{Introduction}
\label{sec:introduction}
Dialogue evaluation is an unsolved challenge in current human-machine interaction research. This is particularly true for open-domain conversation, where compared to task-oriented dialogue (i.e., restaurant reservations), we do not have a finite set of entities and intents, and speakers' goals are not defined a priori.
In this work, we address the problem of dialogue evaluation from the perspective of dialogue \textit{coherence} and how this concept can be formalized and evaluated. Our approach could be applied to both task-oriented and non-task-oriented dialogue.

Coherence in language, i.e., the property which determines that a given text is a logical and consistent whole rather than a random collection of sentences, is a complex multifaced concept which has been defined in different ways and to which several factors contribute \cite{redeker2000coherence}, e.g., rhetorical structure \cite{hobbs1979coherence}, topics discussed, and grounding \cite{traum1994computational}.

While much recent work has focused on coherence for response generation \cite{serban2016generative,li2016diversity, yi2019towards}, we argue that there is still much to be understood regarding the mechanisms and substructures that affect human perception of dialogue coherence. In our approach, in particular, we are interested in studying the patterns of distribution of entities and Dialogue Acts (DAs), in regards to dialogue coherence.

Approaches to coherence based on entities have been studied extensively by the Natural Language Processing literature \cite{joshi1979centered,grosz1995centering}, especially in text (e.g., news, summaries).
Coherence evaluation tasks proposed by this literature \cite{barzilay2008modeling} have the advantage of using weakly supervised training methodologies, but mainly considering documents as-a-whole, rather than evaluating coherence at the utterance level. The dialogue literature \cite{sacks1995lectures,schegloff1968sequencing}, on the other hand, has focused mainly on coherence in connection to DAs, a generalized version of intents in dialogue (e.g., \textit{yes-no-question, acknowledgement}). Recent work \cite{cervone2018coherence}, in particular, showed the importance of both DAs and entities information for coherence modeling in dialogue. However, even in this case dialogue coherence was rated for entire dialogues rather than studying turn coherence structures.

In this work, we investigate underlying conversation turn substructures in terms of DA and entity transitions to predict turn-by-turn coherence in dialogue. 
We start by annotating a corpus of spoken open-domain conversations with turn coherence ratings, the Switchboard Coherence corpus (SWBD-Coh)\footnote{The Switchboard Coherence corpus is available for download at: \url{https://github.com/alecervi/switchboard-coherence-corpus}}, and perform an analysis of the human perception of coherence in regards to DAs and entities. A multiple regression analysis shows the importance of both types of information for human rating of coherence. 
Secondly, we present novel neural models for turn coherence rating that combine DAs and entities and propose to train them using response selection, a weakly supervised methodology. While previous work on response selection \cite{lowe2017training,yoshino2019dialog} is mainly based on using the entire text as input, we deliberately choose to use only entities and DAs as input to our models, in order to investigate entities and DAs as a signal for turn coherence.
Finally, we test our models on the SWBD-Coh dataset to evaluate their ability to predict turn coherence scores \footnote{The code for the models presented in this work can be found at: \url{https://github.com/alecervi/turn-coherence-rating}}.

The main contributions of this work are:
\begin{itemize}[noitemsep]
    \item creating the Switchboard Coherence corpus, a novel human-annotated resource with turn coherence ratings in non-task-oriented open-domain spoken conversation;
    \item investigating human perception of coherence in spoken conversation in relation to entities and DAs and their combination;
    \item proposing novel neural coherence models for dialogue relying on entities and DAs;
    \item exploring response selection as a training task for turn coherence rating in dialogue.
\end{itemize}

\section{Related work}
\label{sec:related}

\textbf{Coherence evaluation in text}
Coherence models trained with weakly supervised methodologies were first proposed for text with applications to the news domain and summarization \cite{barzilay2008modeling}. These models rely on the entity grid, a model that converts the entities (Noun Phrases) mentioned in the text to a sentence-by-sentence document representation in the form of a grid.
The tasks on which coherence models in this line of research are usually evaluated are \textit{sentence ordering} \cite{barzilay2008modeling}, i.e., ranking original documents as more coherent than the same documents with the order of all sentences randomly permuted, and \textit{insertion}, i.e., ranking original documents as more coherent than documents with only one sentence randomly misplaced.
These tasks are still considered standard to this day and found wide applications, especially for text \cite{farag2018neural,clark2018neural}. Recent models proposed for these tasks are based on Convolutional Neural Networks \cite{nguyen2017neural}, also applied to thread reconstruction \cite{joty2018coherence}, while the current State-of-the-art is based on a combination of bidirectional Long Short-Term Memory encoders and convolution-pooling layers \cite{moon2019unified}.
These tasks, however, consider documents as-a-whole and rely mainly on entities information. \\
\textbf{Coherence evaluation in dialogue}
Models for dialogue coherence evaluation have mainly been explored using supervised approaches, i.e., training on corpora with human annotations for coherence, mostly at the turn level \cite{higashinaka2014evaluating,gandhe2016semi,venkateshevaluating, lowe2016evaluation, yi2019towards}.
Different approaches tried to apply the standard coherence tasks to conversational domains such as dialogue and threads, but mainly considering the evaluation of dialogues as-a-whole \cite{purandare2008analyzing,elsner2011disentangling,cervone2018coherence, vakulenko2018measuring,joty2018coherence,mesgar2019neural, zhou2019hierarchical}. 
In particular, \citet{cervone2018coherence} found that discrimination might be over-simplistic for dialogue coherence evaluation when considering Dialogue Act (DA) information. 
In this work, we propose a novel framework to model entities and DAs information for turn coherence prediction using a weakly supervised training methodology. Furthermore, our focus is on predicting coherence of single turns rather than entire dialogues. \\
\textbf{Response Selection}
As a task, response selection has become a standard \cite{lowe2017training,yoshino2019dialog, kumar2019practical} for training both task-oriented and non-task-oriented retrieval-based dialogue models. The task proved to be useful for evaluating models in task-oriented (Ubuntu), social media threads (Twitter Corpus), and movie dialogues (SubTle Corpus) \cite{lowe2016evaluation}. 
Recently the task has also been proposed for pre-training models for task-oriented dialogue \cite{henderson-etal-2019-training} and for Dialogue Act tagging \cite{mehri2019pretraining}.
In this work, we investigate response selection as a task for training coherence rating models for spoken dialogue. Additionally, while response selection models are usually based on the entire text as input \cite{lowe2017training}, we rely solely on entities and DAs information, in order to investigate their effect on turn coherence perception.

\section{Methodology}

In this work, we are interested in the relation between Dialogue Acts (DAs) and entities and how they can be modelled to train automatic predictors of next turn coherence in non-task-based dialogue.

Our hypothesis is that both entities and DAs are useful to predict the coherence of the next turn.
In order to verify such hypothesis, we first perform an analysis of entities and DAs patterns of distribution in the Switchboard Coherence (SWBD-Coh) corpus, a novel dataset of human-human telephone conversations from Switchboard annotated with human coherence ratings per turn.

Secondly, we hypothesize that we can model entities and DAs to predict next turn coherence ratings.
Rather than using supervised data for coherence prediction, we use a weakly supervised training methodology, i.e. training on the task of response selection (which proved useful for other dialogue tasks \cite{henderson-etal-2019-training}) and testing on coherence ratings.
In response selection given a \textit{context}, i.e. the history of the dialogue up to the current turn, and a \textit{list of next turn candidates}, models are asked to rank candidates according to their appropriateness with the previous dialogue history.
The positive training samples for this task are automatically generated by randomly selecting a given turn in a dialogue, and considering this turn as a positive (coherent) example with the current history of the conversation (the context). 
Negative samples are generated by selecting other random dialogue turns, assuming that they will mostly be not appropriate as the next turn in the dialogue.
In particular, we investigate two methodologies to generate negative samples from the training data automatically:\\
\textbf{Internal swap}: a random turn is selected from a subsequent part of the same conversation. We assume this task to be harder for coherence evaluation since typically conversations do not have radical topic shifts. \\
\textbf{External swap}: a random turn is selected from other conversations. We assume this task to be easier given the probable shifts in topic. 

In our first set of experiments, we thus train our models on response selection.
One of the possible shortcomings of the data generation procedure used in response selection, however, is the amount of false negatives.
Although it is assumed that the majority of negative samples generated with this methodology will not be appropriate for the context, there could still be cases in which they are.

In order to verify the performance of our models based on DAs and entities to predict real human coherence judgments, in our second set of experiments models are tested on SWBD-Coh. Analogously to response selection, in turn coherence rating models need to rank next turn candidates given the history of the dialogue. In this case, however, the ranking is not binary but is rather based on a graded coherence rating given by humans for next turn candidates (for further details on the SWBD-Coh corpus see Section \ref{sec:data}).

\begin{figure*}[ht]
  \includegraphics[width=\textwidth]{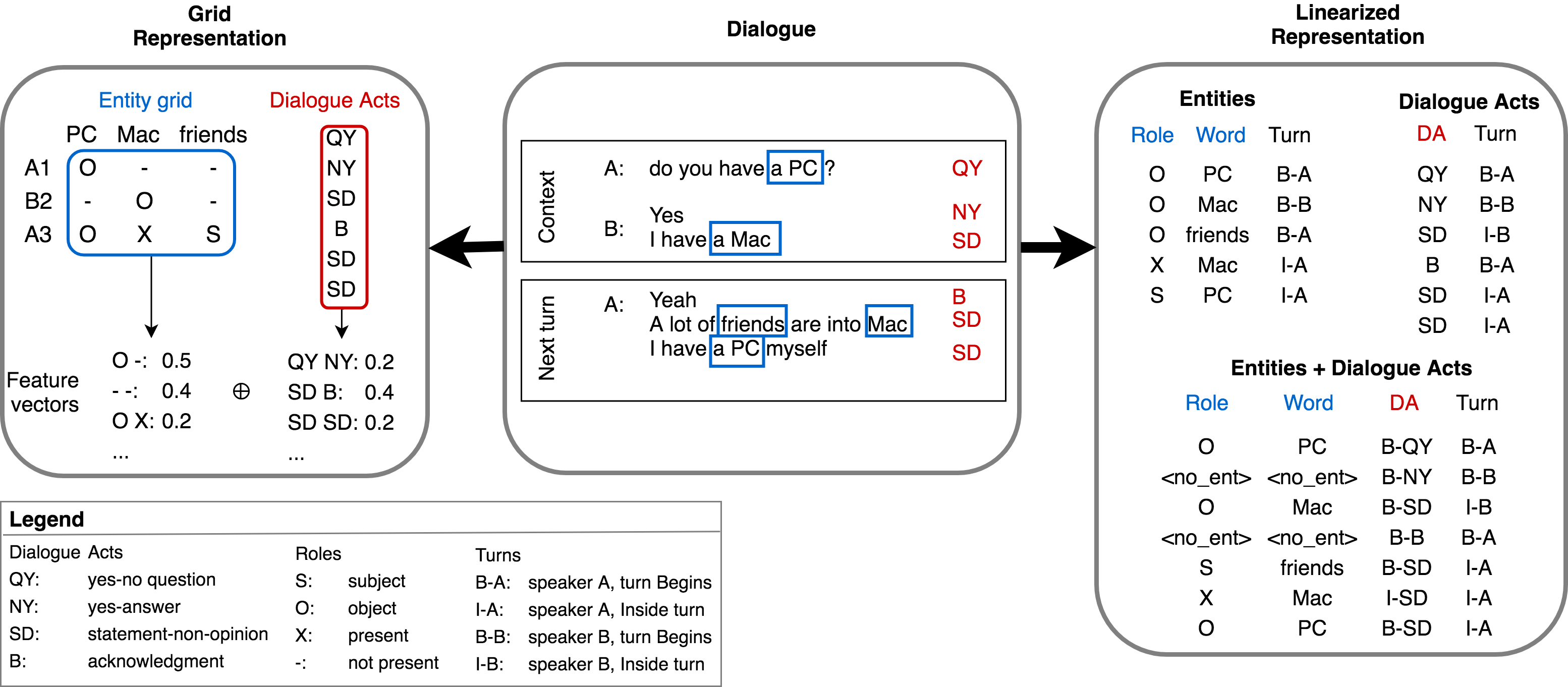}
  \caption{A source \textbf{dialogue} (at the center of the figure) is transformed into a \textbf{grid representation} (left) and into a \textbf{linearized representation} (right). In the grid representation, entities and Dialogue Acts (DAs) are transformed into feature vectors and can then be concatenated. Our linearized representation, i.e. the input to our neural models, shows 3 different possibilities: one where we only consider entity features at the turn level (top-left), another one which considers only DA features (top-right), and a joined one where DAs and entities are combined (bottom).}
  \label{fig:input}
  \vspace{-0.1cm}
\end{figure*}

\section{Data}
\label{sec:data}
\begin{table}
\centering
\scalebox{0.95}{
\begin{tabularx}{\columnwidth}{llll}
\hline
 & \textbf{Train} & \textbf{Dev} & \textbf{Test} \\
\hline
No. source dialogues & 740 & 184  & 231 \\
No. insertion points & 7400 & 1840 & 2310 \\
No. pos/neg pairs & 66600 & 16560 & 20790 \\
\hline
\end{tabularx}
}
\caption{Train, development and test data size for response selection for both Internal and External Swap.}
\label{tbl:datatrain}
\end{table}
The dataset chosen for our experiments is the Switchboard Dialogue Act corpus \cite{stolcke2000dialogue} (SWBD-DA), a subset of Switchboard annotated with DA information. The Switchboard corpus is a collection of human--human dyadic telephone conversations where speakers were asked to discuss a given topic.
This dataset was chosen both to ensure comparability with previous work on dialogue coherence and because it is open-domain. Also, this corpus has DA annotations. Interestingly, SWBD-DA is a real-world (transcribed) spoken corpus, so we have sudden topic changes, overlap speech, disfluencies and other typical characteristics of spoken interaction. Since our goal was to study coherence in a real-world spoken dialogue setting, rather than removing these features as errors, we considered them an integral part of spoken conversations and did not remove them. 

\paragraph{Response Selection}
Source dialogues are split into train, validation, and test sets (see Table \ref{tbl:datatrain}) using the same distribution as \citet{cervone2018coherence}. 
For each dialogue, we randomly choose ten insertion points. Each insertion point is composed by a context (dialogue history up to that point) and the original turn following that context (regarded as positive).
In order to have 10 next turn candidates, for each insertion point 9 adversarial turns (regarded as negatives) are then randomly selected either from subsequent parts of the dialogue, i.e. Internal Swap (IS), or from other dialogues, i.e. External Swap (ES), within the same data subset, so that for example external adversarial turns for training are only taken from other source dialogues in the training set.

\paragraph{Switchboard Coherence corpus}
The dataset for turn coherence rating, the Switchboard Coherence corpus (SWBD-Coh), was created using as source dialogues the ones from SWBD-DA which are in the testset of \citet{cervone2018coherence}. The data were annotated using Amazon Mechanical Turk (AMT). 1000 insertion points were randomly selected, following the constraints that the context (dialogue history up to the original turn) could be between 1 and 10 turns length. 
Since in this task we want to evaluate the coherence of a given turn with the previous dialogue history, 1 turn of context was the minimum required. We set the maximum length to 10 turns to reduce annotation time.
For each insertion point, six adversarial turns were randomly selected, besides the original one (3 using the IS methodology, 3 using the ES one) for a total of 7 turn candidates. Overall the SWBD-Coh dataset is thus composed of 7000 pairs (1000 contexts $\times$ 7 turns).
\\
Each context and turns pair was annotated by 5 AMT workers with coherence ratings. More specifically, for each dialogue workers were presented with the dialogue history up to the insertion point and the next turn candidates (randomly shuffled). Workers were asked to rate on a scale of 1 (not coherent), 2 (not sure it fits) to 3 (coherent) how much each response makes sense as the next natural turn in the dialogue. All workers (37) who annotated the dataset were first evaluated on a common subset of 5 dialogues where they had an average Weighted Kappa agreement with quadratic weights with two gold (internal) annotators of $\kappa=0.659$ (min: 0.425, max: 0.809, STD: 0.101) and among each other an average leave-one-out correlation of $\rho=0.78$ (i.e. correlating the scores of each worker with mean scores of all other workers who annotated the same data), following the approach used in other coherence rating datasets \citep{barzilay2008modeling, lapata2006automatic}. \footnote{More details about our data collection procedure are available in Appendix \ref{sec:appendixDataCollection}.}
Scores for each candidate turn were then averaged across all annotators. Original turns were regarded on average as more coherent ($\mu=2.6$, SD$=0.5$) than adversarial turns, while turns generated with IS were considered more coherent ($\mu=1.8$, SD$=0.7$) than the ones generated via ES ($\mu=1.4$, SD$=0.6$). 

\section{Data analysis}
\label{sec:data_analysis}

In this section, we analyse the Switchboard Coherence (SWBD-Coh) dataset in regards to the distribution of Dialogue Acts (DAs) and entities. In particular, we are interested in analysing which features might affect human judgement of coherence of a given next turn candidate. For entities, we analyse two features: the number of entities mentioned in the next turn candidate that overlap with entities introduced in the context and the number of novel entities introduced in the turn. Additionally, we create a binary feature for each DA type that registers the presence of that DA in the turn candidate. 

We use multiple regression analysis to verify how these different features correlate with human coherence ratings. Table \ref{tbl:dataAnalysisMCC}, reports the Multiple Correlation Coefficient (MCC) of regression models using R squared and Adjusted R squared \cite{theil1961economic}, adjusted for the bias from the number of predictors compared to the sample size.
The results of our analysis indicate that the best MCC, 0.41 when calculated with the Adjusted R squared, is achieved when combining all features, both from entities and DAs. 
Moreover, in the lower part of Table \ref{tbl:dataAnalysisMCC} we report some of the features that proved to be the most relevant for predicting human coherence ratings.
In general, it seems that while the entities overlapping the previous context seems to affect positively human coherence judgements, the DAs that most affect ratings do so in a negative way and seem to be mostly contentful DAs, such as \textit{statement-opinion}, rather than DAs which typically present no entities, such as \textit{acknowledge}. Our interpretation is that, in cases when there are no overlapping entities with the context, these DAs might signal explicit examples of incoherence by introducing unrelated entities.


\begin{table}
\centering
\setlength{\tabcolsep}{2.5pt}
\scalebox{0.95}{
\begin{tabularx}{\columnwidth}{lll}
\hline
 & \heading{\textbf{MCC$R^{2}$}} & \heading{\textbf{MCC$AR^{2}$}} \\
\hline
Entities  & \heading{0.27} & \heading{0.26}  \\
DAs & \heading{0.34} & \heading{0.29} \\
All (Entities + DAs) & \heading{\textbf{0.45}} & \heading{\textbf{0.41}} \\
\hline\hline
\textit{Relevant features in All} & \heading{\textit{Coeff.}} & \heading{\textit{Sign.}} \\
 \hline
Overlapping entities & \heading{0.26} & \heading{**} \\
DA: decl. yes-no-question & \heading{-0.48}  & \heading{*} \\
DA: statement-opinion & \heading{-0.31} & \heading{**} \\
DA: statement-non-opinion & \heading{-0.30} & \heading{**} \\
DA: acknowledge & \heading{0.27} & \heading{**} \\
\hline
\end{tabularx}
}
\caption{Multiple Correlation Coefficients (MCC) from R squared ($R^{2}$) and Adjusted R squared ($AR^{2}$) of different multiple regression models that predict human coherence ratings for candidate turns given a dialogue context (turn coherence rating task) on the Switchboard Coherence corpus. Additionally, we report coefficients and significance (where * denotes $.05\geq$$p$$\geq.01$ and ** $p<.01$) of some relevant features for the best-performing model (All).}
\label{tbl:dataAnalysisMCC}
\end{table}
\section{Models}
We model dialogue coherence by focusing on two features that have been closely associated to coherence in previous literature: the entities mentioned and the speakers' intents, modelled as Dialogue Acts (DAs), in a conversation.
Our models explore both the respective roles of entities and DAs and their combination to predict dialogue coherence. 
We investigate both standard coherence models based on Support Vector Machines (SVM) and propose novel neural ones.

\subsection{SVM models}
The entity grid model \cite{barzilay2008modeling} relies on the assumption that transitions from one syntactic role to another of the same entities across different sentences of a text indicate local coherence patterns. This assumption is formalized by representing a text (in our case, a dialogue) as a grid, as shown in Figure \ref{fig:input}. For each turn of the dialogue we extract the entities, i.e. Noun Phrases (NPs), and their respective grammatical roles, i.e. whether the entity in that turn is subject ($S$), direct object ($O$), neither ($X$), or it is not present ($-$).
Each row of the grid represents a turn in the dialogue, while each column represents one entity (in Figure \ref{fig:input}, for example, the first turn of speaker A is represented by the first row of the grid $O--$). Using this representation, we can derive feature vectors to be used as input for Machine Learning models by extracting probabilities of all role transitions for each column. 

More formally, the coherence score of a dialogue $D$ in the entity grid approach can be modelled as a probability distribution over transition sequences for each entity $e$ from one grammatical role $r$ to another for all turns $t$ up to a given history $h$ (see Eq. 4 in \citet{lapata2005automatic}):

%
\begin{equation}
\small
p_{cohEnt}(D) \approx \frac{1}{m\cdot n} \prod_{e=1}^{m}\prod_{t=1}^{n} 
p(r_{t,e}|r_{(t-h),e}...r_{(t-1),e})
\end{equation}

The probabilities for each column (entity) are normalized by the column length $n$ (number of turns in the dialogue) and the ones for the entire dialogue by the number of rows $m$ (number of entities in the dialogue). In this way, we obtain the feature vectors shown in Figure \ref{fig:input} where each possible roles transition of a predefined length (e.g. $O-$) is associated with a probability. These feature vectors are then given as input to a Support Vector Machine (SVM) in the original model.

Following \citet{cervone2018coherence}, we can use the same approach to construct similar feature vectors for DAs information:
\begin{equation}
\small
p_{cohDA}(D) \approx \frac{1}{n} \prod_{i=1}^{n}p(d_{i}|d_{(i-h)}...d_{(i-1)})
\end{equation}
Here the coherence score of a dialogue is given by the probability of the entire sequence of DAs ($d$) for the whole dialogue, normalized by column length ($n$), i.e. the number of DAs for each turn. 

The joint model, the one combining entity and DA information, concatenates feature vectors obtained from both. While other ways of combining DA and entities have been explored in \citet{cervone2018coherence}, the authors report that practically a concatenation resulted in the best performances across all tasks, probably due to data sparsity issues.

Indeed among the limitations of the entity grid, there is data sparsity: for example for an entity appearing only in the last turn of a dialogue we need to add a column to the grid which will be mostly containing ``empty'' $--$ transitions (see \textit{friends} in Figure \ref{fig:input}).
Another problem of this approach is the fact that the model is not lexicalized since we only keep role transitions when computing the feature vectors for the entities. Furthermore, the model makes the simplifying assumption that columns, thus entities, are independent from each other. 
\subsection{Neural models}
Our neural coherence models for dialogue are based on bidirectional Gated Recurrent Units (biGRU).
While other neural coherence models \cite{nguyen2017neural,joty2018coherence} rely directly on the grid representation from \citet{barzilay2008modeling}, we explore a novel way to encode the dialogue structure. 
The input to our biGRUs is a sequential representation of the dialogue.

\subsubsection{Sequential input representation}

We linearize the structure of a dialogue composed by entities, DAs and turns into flat representations for our neural models, as in Figure \ref{fig:input}. These representations can then be mapped to an embedding layer and joined via concatenation. 
We consider three cases: (i) the case in which we model entity features; (ii) the one in which we consider DAs information; (iii) the one in which we combine both. 

\paragraph{Entities encodings}
In our approach, entities are Noun Phrases, as in the entity grid approach. 
For each dialogue, we consider the sequence of entities ordered according to their appearance in the conversation (see Figure \ref{fig:input}). 
Entities are represented either by their grammatical roles \textit{ent\textsubscript{\textit{role}}} in the dialogue (using the same role vocabulary $V_{r}$ of the original grid), their corresponding words \textit{ent\textsubscript{\textit{word}}} (from a vocabulary $V_{w}$), or by both.
Another feature which can be added to this representation is the \textit{turn} (whether A or B is talking). 
This feature could be useful to encode the dyadic structure of the dialogue and how this might be related to entity mentions. In order to better encode the boundaries of speaker turns, turns are mapped to the IOB2 format (where the Outside token is removed because naturally never used for turns), for a resulting turn vocabulary $V_{t}$ size of 4 tags (2 speakers x 2 IOB tags used). Special tokens ($<$no\_ent$>$) are added to both $V_{w}$ and $V_{r}$ for cases in which turns do not present any entities.

\paragraph{DAs encodings}
In case we consider only \textit{DAs} features, our input representation becomes a sequence of DAs for the whole dialogue history so far, drawn from a vocabulary $V_{d}$. Also, in this case, \textit{turn} features can be added to mark the turn-wise structure of the DA sequence, using the same vocabulary $V_{t}$ previously described.

\paragraph{Entities + DAs encodings}
We combine entities and DAs by considering the sequence of entities in order of their appearance within each DA and encoding DAs into IOB2 format, as previously done for turn features. In this setting, thus, the vocabulary $V_{d}$ has double the size, compared to the setting where we consider only DAs. Analogously to previous settings, turn features can be added to encode turn boundaries. 

It can be noticed how our representation is less sparse compared to both the original grid \cite{barzilay2008modeling} and recently proposed models \cite{nguyen2017neural}, which take as input grid columns directly. Furthermore, compared to the original grid, our representation is lexicalized.
\subsubsection{Architecture}
\begin{figure}[ht]
  \includegraphics[width=6.9cm]{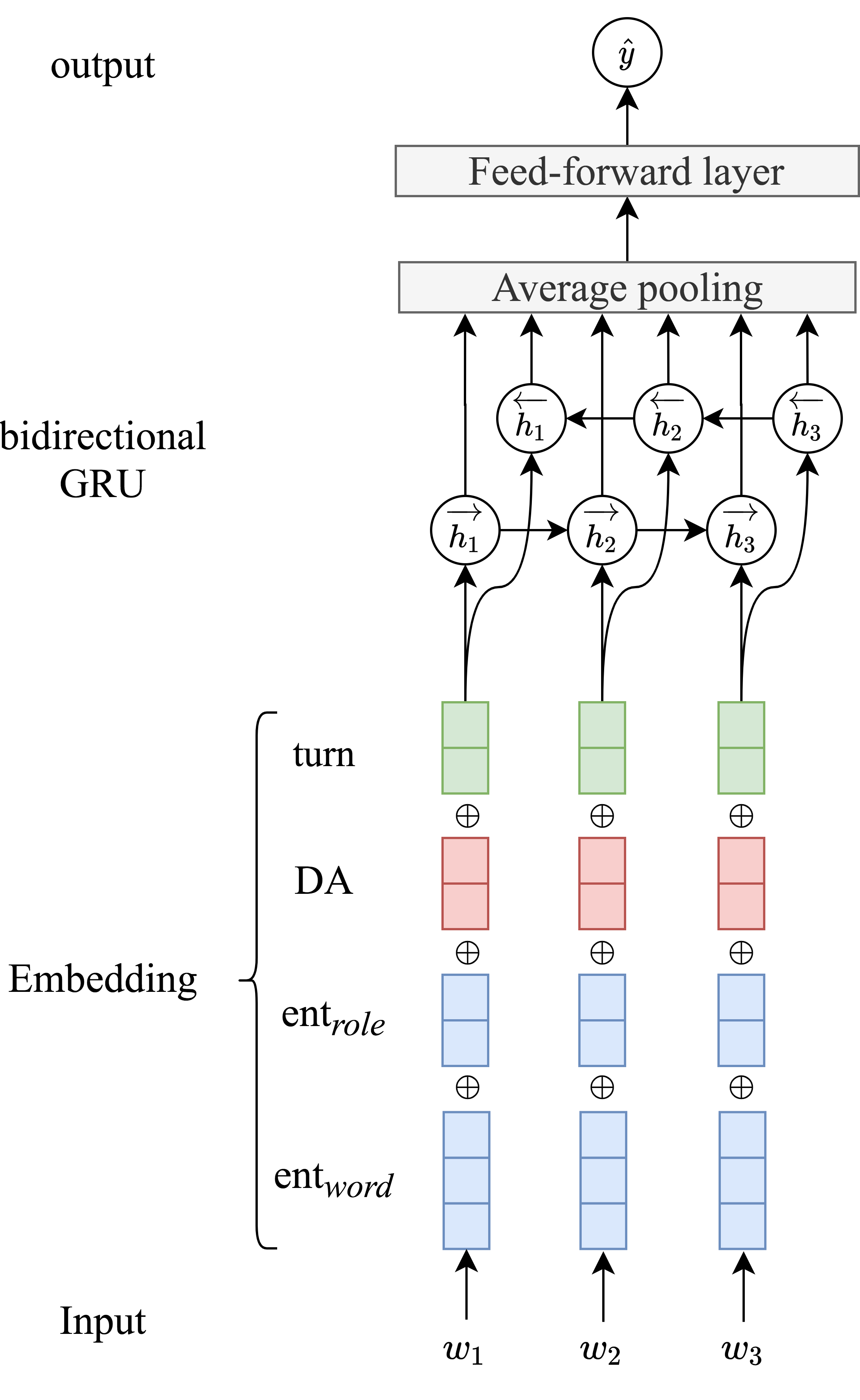}
  \caption{Our proposed architecture based on bidirectional GRUs with input entity word embedding (\textit{ent\textsubscript{\textit{word}}}) and grammatical role (\textit{ent\textsubscript{\textit{role}}}), Dialogue Act (\textit{DA}) and speaker \textit{turn} features.}
  \label{fig:architecture}
\end{figure}
The architecture of our models is shown in Figure \ref{fig:architecture}.
In the first layer of the network each input feature (\textit{ent\textsubscript{\textit{role}}}, \textit{ent\textsubscript{\textit{word}}}, \textit{DA}, \textit{turn}) is mapped to a $d$-dimensional dense vector by looking up into their respective embedding matrix $\textbf{E}$, one per feature type. All features vectors obtained can then be combined using concatenation.
This vector is then recursively passed to the bidirectional GRU layers and then to a mean pooling layer. Finally, the output is passed through a feed-forward neural network with one hidden layer and ReLU as non-linearity.

Our models are trained using a Margin-ranking loss with a margin of 0.5 using the following equation:
\begin{equation}
    \textup{loss}(x,y) = \textup{max}(0, -y*(x1-x2) + \textup{margin})
\end{equation}
where $x1$ and $x2$ are respectively the original dialogue and the adversarial one and $y=1$.
In this way, the model is asked to rank the original dialogue higher (more coherent) than the adversarial one. 
The model is trained by Stochastic Gradient Descent, using the Adam update rule \cite{DBLP:journals/corr/KingmaB14}.
\section{Experimental set-up}
\label{sec:expsetup}
\textbf{Preprocessing} Entities, i.e. Noun Phrases (NPs), and their syntactic roles were extracted and preprocessed with \citet{cervone2018coherence}'s pipeline \footnote{\url{https://github.com/alecervi/Coherence-models-for-dialogue}}. Following the original entity grid formulation \cite{barzilay2008modeling}, only NPs heads were kept. The DAs are taken from annotations on SWBD-DA (using the standard reduction to 42 tags compared to the DAMSL ones). 

\paragraph{Evaluation} 
For evaluating response selection, we use pairwise Accuracy, the metric used in standard coherence tasks, which evaluates the ability of the model to rank original turns higher than each adversarial one. However, this metric is not indicative of the global ranking of all candidate turns for a given context. For this reason, we add two ranking metrics to evaluate our models: Mean Reciprocal Rank (MRR), which evaluates the average of reciprocal ranks of all candidate turns for a context, and Recall at One (R@1) and Two (R@2), also used in previous work on response selection \cite{lowe2017training,zhou2018multi} to assess the ability of the model to rank original turns respectively within the first or second rank among all candidates. \\
Compared to response selection, where we have a binary choice between coherent and negative turns, in turn coherence rating, we have a set of candidate turns each associated to a coherence score. In this case, we use Accuracy, MRR, R@1 and Normalized Discounted Cumulative Gain (nDCG) to evaluate our models. Accuracy was computed only for cases in which the rating of the turn was not identical across two candidate turns. MRR and R@1 were computed dynamically, that is considering the turn with the highest score within that particular context as the best one in the rank. 
The nDCG metric \cite{jarvelin2002cumulated} assesses the gain of a candidate according to its rank among all candidates. Compared to previous metrics, nDCG allows taking into account the relevance (in our case, the coherence score) of candidates. 
For all metrics considered, if our models predicts the same score for two candidates, we always assume models made a mistake, i.e. among candidates with the same predicted score positive examples are ranked after the negative ones. 
\begin{table*}
\setlength{\tabcolsep}{5pt}
\centering
\scalebox{0.94}{
\begin{tabularx}{\textwidth}{lrrrr|rrrr}
\cline{2-9}
& \multicolumn{4}{c}{\textbf{Internal Swap}}
& \multicolumn{4}{c}{\textbf{External Swap}}
\\
\cline{2-9}
& \textbf{Acc.}
& \textbf{MRR}
& \textbf{R@1}
& \textbf{R@2}
& \textbf{Acc.}
& \textbf{MRR}
& \textbf{R@1}
& \textbf{R@2}\\
\hline
Random
& 50.0 & 0.293 & 0.099 & 0.198 
& 50.0 & 0.293 & 0.099 & 0.198 \\
\hline
\hline
SVM ent\textsubscript{\textit{role}} (Entity Grid)
& 36.6 & 0.260 & 0.103 & 0.178
& 39.5 & 0.246 & 0.096 & 0.126 \\
SVM DA \cite{cervone2018coherence}
& 60.6 & 0.398 & 0.206 & 0.335
& 61.3 & 0.403 & 0.212 & 0.346 \\
SVM ent\textsubscript{\textit{role}} + DA \cite{cervone2018coherence}
& 62.7 & 0.417 & 0.222 & 0.365
& 64.3 & 0.437 & 0.251 & 0.380 \\
\hline
\hline
biGRU ent\textsubscript{\textit{role}} 
& 41.8 & 0.294 & 0.120 & 0.217 
& 45.5 & 0.293 & 0.117 & 0.210 \\
biGRU ent\textsubscript{\textit{role}} + turn 
& 43.3 & 0.295 & 0.120 & 0.214 %
& 45.9 & 0.293 & 0.115 & 0.211 \\
biGRU ent\textsubscript{\textit{word}}
& 47.8 & 0.324 & 0.151 & 0.252 %
& 56.4 & 0.397 & 0.236 & 0.337 \\
biGRU ent\textsubscript{\textit{word}} + turn 
& 49.0 & 0.331 & 0.162 & 0.255 %
& 56.9 & 0.400 & 0.241 & 0.341 \\
biGRU ent\textsubscript{\textit{role}} + ent\textsubscript{\textit{word}} + turn 
& 48.6 & 0.327 & 0.156 & 0.253 %
& 56.1 & 0.394 & 0.232 & 0.338 \\
\hline
biGRU DA 
& 72.4 & 0.484 & 0.276 & 0.443 %
& 72.6 & 0.486 & 0.278 & 0.447 \\
biGRU DA + turn
& 74.0 & 0.501 & 0.297 & 0.464 %
& 74.1 & 0.508 & 0.305 & 0.475 \\
\hline
biGRU ent\textsubscript{\textit{word}} + DA + turn
& \textbf{75.1} & 0.520 & \textbf{0.321} & 0.484 %
& \textbf{77.3} & \textbf{0.550} & \textbf{0.355} & \textbf{0.530} \\
biGRU all
& 75.0 & \textbf{0.521} & \textbf{0.321} & \textbf{0.489} %
& 77.2 & 0.549 & 0.354 & 0.529 \\
\hline
\end{tabularx}
}
\caption{Average (5 runs) of Accuracy (Acc.), Mean Reciprocal Rank (MRR) and Recall at one (R@1) and two (R@2) for response selection using both data generation methodologies (Internal and External Swap) on Switchboard. 
}
\label{tbl:pred}
\end{table*}

\paragraph{Models' settings} 
Grid models, based on SVMs, were trained with default parameters using SVM\textsuperscript{light} preference kernel \cite{joachims2002optimizing}) as in the original model \cite{barzilay2008modeling}. 
For saliency, i.e. the possibility of filtering entities according to their frequency, and transitions length we follow the default original grid parameters (saliency:1, transitions length:2). 
For neural models, implemented in Pytorch \cite{paszke2019pytorch}, parameters were kept the same across all models to ensure comparability. The learning rate was set to 0.0005, batch size to 32, with two hidden biGRU layers of size 512.
Embedding sizes for all features were set to 50--dimensions, except for word embeddings which had dimension 300. Models run for a maximum of 30 epochs with early stopping, based on the best MRR score on the development set.

\section{Results}
\label{sec:results}
\begin{table}
\setlength{\tabcolsep}{1.9pt}
\scalebox{0.92}{
\begin{tabularx}{\columnwidth}{lrrrrrr}
\cline{2-6}
& \textbf{Train}
& \textbf{Acc.}
& \textbf{MRR}
& \textbf{R@1}
& \textbf{nDCG}
\\
\cline{1-6}
Random & 
& 50.0 & 0.479 & 0.234 & 0.645 \\
\cline{1-6}
biGRU & IS
& 42.7 & 0.395 & 0.174 & 0.621 \\
ent\textsubscript{\textit{word}} + turn & ES
& 50.4 & 0.444 & 0.229 & 0.679 \\
\cline{1-6}
biGRU & IS
& 56.0 & 0.553 & 0.326 & 0.717 \\
DA + turn & ES
& 56.0 & 0.558 & 0.337 & 0.725 \\
\cline{1-6}
biGRU & IS
& 58.5 & 0.575 & 0.358 & 0.738 \\
ent\textsubscript{\textit{word}} + DA + turn & ES
& \textbf{61.1} & \textbf{0.583} & \textbf{0.369} & \textbf{0.760} \\
\cline{1-6}
\end{tabularx}
}
\caption{Average (5 runs) of Accuracy (Acc.), Mean Reciprocal Rank (MRR), Recall at one (R@1) and Normalized Discounted Cumulative Gain (nDCG) for turn coherence rating for models trained using either Internal (IS) or External Swap (ES) on the Switchboard Coherence corpus.}
\label{tbl:humanresult}
\end{table}
In this section, we report the results of our models for response selection. The best performing models on response selection are then evaluated on the turn coherence rating task using the Switchboard Coherence (SWBD-Coh) corpus as testset. For both tasks we compare our models to a random baseline. All reported results for neural models are averaged across 5 runs with different seeds.

\paragraph{Response selection}
The results for response selection are reported in Table \ref{tbl:pred}.
Neural models seem to capture better turn-level coherence compared to classic grid SVM-based approaches. In both data generation methodologies, Internal (IS) and External Swap (ES), SVM coherence models are outperformed by neural ones for all metrics considered. 
As expected, entity features (ent\textsubscript{\textit{role}}, ent\textsubscript{\textit{word}}) play a more prominent role in ES compared to IS.
In both cases, entity features seem to be better captured by neural models relying on our proposed input representation.
When considering lexical information (ent\textsubscript{\textit{word}}), however, ent\textsubscript{\textit{role}} features seem less relevant. This might be due to the fact that spoken dialogue has usually less complex syntactic structures compared to written text. Furthermore, parsers are usually trained on written text, and thus might be more error-prone when applied to dialogue where there are disfluencies, sudden changes of topics, etc.
We notice that DAs alone (without entity information) play an important role in both IS and ES.
Turn features capturing speaker information seem helpful for both DAs and entities.
\\
In general, the combination of DAs and entities gives the best results both in SVM and neural models for both tasks, with the best performing one being the model combining ent\textsubscript{\textit{word}}, DA and turn features and without ent\textsubscript{\textit{role}}.
Additionally, if we compare the IS setting to ES in terms of best MRR, Accuracy and Recall, the former seems more difficult. This confirms our expectations that IS might be an harder task for coherence.

\paragraph{Turn coherence rating}
A selection of best performing models for entities, DAs and their combination were tested on the SWBD-Coh dataset. Table \ref{tbl:humanresult} shows models' results under both training conditions, i.e. either using IS or ES data. 
The lowest performing model seems to be the one based solely on entity features (ent\textsubscript{\textit{word}} $+$ turn), while models combining DA with entities information (ent\textsubscript{\textit{word}} $+$ DA $+$ turn) are the best performing ones. Additionally, models trained on ES data perform better than those trained on IS across all conditions.

\section{Conclusions}
\label{sec:conclusions}
In this work, we investigate how entities and Dialogue Acts (DAs) are related to human perception of turn coherence in dialogue. In order to do so, we create a novel dataset, the Switchboard Coherence (SWBD-Coh) corpus, of transcribed open-domain spoken dialogues annotated with turn coherence ratings. A statistical analysis of the corpus confirms how both entities and DAs affect human judgements of turn coherence in dialogue, especially when combined.
Motivated by these findings, we experiment with different models relying on entities and DAs to automatically predict turn coherence, i.e. standard coherence models and novel neural ones. In particular, we propose a less sparse alternative, compared to the entity grid, to encode entities and DAs information.
Rather than using data annotated explicitly for the task, i.e. coherence prediction, we explore two response selection methodologies for training.  
We find that our newly proposed architecture outperforms standard ones in response selection.
Finally, we test our models on the SWBD-Coh corpus in order to evaluate their ability to predict real human turn coherence ratings. Crucially, we find that the combination of DAs and entities gives the best performances. 

For the future work, it would be interesting to investigate how to apply large pretrained models to our task, such as BERT \cite{devlin2019bert}. While pretrained models have recently been successfully explored for text-based response selection \cite{kim2019eighth, henderson2019training}, integrating them with our proposed input representation is not a straightforward task since such models typically rely on the whole textual context, while our models do not. 

While there is still much to understand regarding turn coherence in dialogue, we believe our work could be a first step towards uncovering the relation between DAs and entities in open-domain spoken dialogue. Moreover, we believe that the SWBD-Coh corpus could become a useful resource for the community to study coherence in open-domain spoken dialogue.

\section*{Acknowledgments}
The research leading to these results has received funding from the European Union – H2020 Programme under grant agreement 826266: COADAPT.


\bibliography{anthology,acl2020}
\bibliographystyle{acl_natbib}

\newpage
\newpage
\appendix



\section{Appendix A: Switchboard Coherence corpus data collection procedure}
\label{sec:appendixDataCollection}

Coherence rating is an inherently subjective task and could be challenging especially for a dataset of transcribed real-world open-domain human-human conversation like Switchboard, where we have possible interruptions, overlaps and disfluencies naturally occurring.
Hence, in order to ensure we collected reliable judgements for turn coherence, we followed a multi-step procedure to build the Switchboard Coherence (SWBD-Coh) corpus using Amazon Mechanical Turk (AMT).

\subsection{Experiment with internal annotators}
First we performed a small-scale annotation experiment to evaluate the feasibility of the task. Two internal annotators, both with Linguistics education, were asked to rate a set of 150 different dialogues randomly selected from the testset from \cite{cervone2018coherence}. The 150 annotation pairs (context + set of candidate turns) were generated using the same procedure described in Section 4 of the paper. The coherence scale was divided into 1 (not coherent), 2 (not sure it fits) and 3 (coherent). Since we wanted to capture a general perception of coherence, rather than bias annotators towards our own intuitions, in the guidelines annotators the task was described as: ``Your task is to rate each candidate on a scale of how much it is \textit{coherent} with the previous dialogue context, that is \textit{how much that response makes sense as the next natural turn in the dialogue}''. 

Since in this case we only have two annotators, we were able to measure their inter-annotator agreement using a weighted kappa score with quadratic weights (since our categories are ordinal). The inter-annotator agreement was of 0.657 (which can be regarded as substantial \cite{viera2005understanding}).
Then, we averaged scores for each candidate turn from both annotators. As shown in Table \ref{tbl:humanEvalDetails}, original turns had higher coherence scores ($\mu=2.66$) compared to adversarial turns, while turns generated with Internal Swap were considered more coherent ($\mu=1.78$) than the ones generated via External Swap ($\mu=1.45$).

\begin{table}
\setlength{\tabcolsep}{3.4pt}
\centering
\scalebox{0.91}{
\begin{tabularx}{\columnwidth}{llll}
\hline
& \textbf{Orig} & \textbf{IS} & \textbf{ES} \\
\hline
$\mu$ score 150 & 2.7 (0.5) & 1.8 (0.7) & 1.4 (0.7) \\
$\mu$ score SWBD-Coh & 2.6 (0.5) & 1.8 (0.7) & 1.4 (0.6) \\
\hline
\end{tabularx}
}
\caption{Comparison of human annotation results for the experiment with two internal annotators (150 dialogues) and the Switchboard Coherence (SWBD-Coh) dataset. Mean scores (and standard deviation) are reported for each candidates group: originals (Orig), internal swap (IS) and external swap (ES).}
\label{tbl:humanEvalDetails}
\end{table}

\subsection{Experiment with AMT}
\begin{table*}
\setlength{\tabcolsep}{2.1pt}
\scalebox{0.74}{
\begin{tabularx}{\textwidth}{lclccc}
 & \multicolumn{1}{l}{}  &  & \multicolumn{3}{c}{\textbf{Models ranks}} \\ 
\cline{1-6}
\multicolumn{1}{c}{\textbf{Context}}                                      & \multicolumn{1}{c}{\textbf{Score}} & \multicolumn{1}{c}{\textbf{Candidates}}                                     & \multicolumn{1}{c}{\textbf{Ent}} & \multicolumn{1}{c}{\textbf{DA}} & \multicolumn{1}{c}{\textbf{Ent+DA}} \\ \cline{1-6} \cline{1-6}
\multicolumn{1}{l|}{\multirow{7}{*}{\begin{tabular}[c]{@{}l@{}}\textit{A}: Okay.\\ \textit{B}: Well, if you are from \textbf{Iowa}, \\ you must be very artsy crafty. \\ Everyone I've ever known from the \textbf{Midwest}\\  can do everything with their \textbf{hands}.\\ \textit{A}: Oh, well, actually I'm from \textbf{California} \\ and before then I was from \textbf{Utah}. So.\end{tabular}}} & \multicolumn{1}{c}{3.0}            & \multicolumn{1}{l}{I didn't know anyone ever moved from \textbf{California} to \textbf{Iowa}?}   & \multicolumn{1}{c}{1}   & \multicolumn{1}{c}{4}           & \multicolumn{1}{c}{1}               \\ 
\cline{2-6} 
\multicolumn{1}{l|}{}  & \multicolumn{1}{c}{2.6}            & \multicolumn{1}{l}{\begin{tabular}[c]{@{}l@{}}Anyway, we are supposed to be talking about \textbf{crafts}. Do you, um, \\ do you have any \textbf{hobbies} that, that you do things with your \textbf{hands}\end{tabular}} & \multicolumn{1}{c}{2}            & \multicolumn{1}{c}{2}           & \multicolumn{1}{c}{2}               \\ \cline{2-6} 
\multicolumn{1}{l|}{}  & \multicolumn{1}{c}{2.2}            & \multicolumn{1}{l}{Right.}   & \multicolumn{1}{c}{4}            & \multicolumn{1}{c}{3}           & \multicolumn{1}{c}{3}               \\ \cline{2-6} 
\multicolumn{1}{l|}{}  & \multicolumn{1}{c}{2.2}            & \multicolumn{1}{l}{Uh-huh.}  & \multicolumn{1}{c}{4}            & \multicolumn{1}{c}{3}   & \multicolumn{1}{c}{3}               \\ 
\cline{2-6} 
\multicolumn{1}{l|}{} & \multicolumn{1}{c}{2.0}            & \multicolumn{1}{l}{Oh, sure.}  & \multicolumn{1}{c}{4}   & \multicolumn{1}{c}{3}  & \multicolumn{1}{c}{3}  \\ 
\cline{2-6} 
\multicolumn{1}{l|}{}  & \multicolumn{1}{c}{1.2}            & \multicolumn{1}{l}{\textbf{bags} some, their most recent, uh, \textbf{needle craft}}          & \multicolumn{1}{c}{3}  & \multicolumn{1}{c}{4}           & \multicolumn{1}{c}{4}               \\ 
\cline{2-6} 
\multicolumn{1}{l|}{}  & \multicolumn{1}{c}{1.0}              & \multicolumn{1}{l}{at least at the end.}  & \multicolumn{1}{c}{5}            & \multicolumn{1}{c}{1}           & \multicolumn{1}{c}{5}               \\ 
\cline{1-6}
\end{tabularx}
}
\caption{Example of how different models relying only on entities (biGRU ent\textsubscript{\textit{word}} + turn), only on DAs (biGRU DA + turn) or both (biGRU ent\textsubscript{\textit{word}} + DA + turn) rank the same group of candidates for a given context. 
}
\label{tbl:modelsRankExample}
\end{table*}
After having assessed the feasibility of the task, we then proceeded to set up the data collection procedure on AMT. 

In order to select workers for our coherence annotation task we first set up a qualification task on AMT. The qualification task consisted of 5 dialogues (taken from the 150 internally annotated) with 7 turn candidates using the same coherence rating scale as in the gold annotation. In order to pass the qualification task a worker had to have a weighted kappa score higher than 0.4 with both our gold annotators. This threshold was decided empirically by first running a small scale experiment with other 4 internal annotators on the qualification task. 37 workers passed the qualification task. The average weighted kappa agreement with the two gold annotators was 0.659 (min: 0.425, max: 0.809, STD: 0.101). In order to calculate the agreement among all the 37 workers on this batch we employ leave-one-out resampling. For each worker who annotated the data we calculate the correlation of her/his scores with the mean ones of all other annotators in the batch. This is repeated for all workers and then averaged. This technique has been used in other coherence annotation experiments \cite{barzilay2008modeling,lapata2005automatic}.

Workers who passed the qualification test could then proceed to annotate the SWBD-Coh data. The data, consisting of 1000 dialogues, was divided into 100 batches of 10 dialogues each. Each batch was annotated by at least 5 workers. In order to remove possible workers who did not perform well on a given batch, we employed a combination of techniques including leave-one-out resampling and average scores given to original turns.
The average leave-one-out correlation per batch for turn coherence rating achieved with this data collection procedure was: $\rho=$0.723 (min: 0.580, max: 0.835, STD: 0.055). 
Interestingly, as shown in Table \ref{tbl:humanEvalDetails}, the average scores per candidate group (original, Internal swap, External swap) match closely the ones obtained in our gold 150 annotation data.

\section{Appendix B Models output example}
\label{sec:appendix_output_example}

Table \ref{tbl:modelsRankExample} shows an example of the ranking given by different models to the same context-candidates pairs in the SWBD-Coh corpus, compared to the average coherence score given by annotators. In particular, we report the ranking given by a model based solely on entities information (biGRU ent\textsubscript{\textit{word}} + turn), another one considering only DAs (biGRU DA + turn) and a third one considering both types of information (biGRU ent\textsubscript{\textit{word}} + DA + turn). All models were trained on response selection using the External Swap methodology. The models output is reported in terms of position in the rank.
Entities appearing in the text are highlighted in bold.

In this example we notice entities overlap information with the previous context proves rather important in order to rank candidates according to coherence.
For example, to rank the candidate with the highest coherence as the first one (\textit{I didn't know anyone ever moved from California to Iowa?}) information regarding the overlapping entities \textit{California} and \textit{Iowa} allows the models encoding entities information to assign the correct rank, while the model relying only on DAs gives the candidate the fourth position in the rank.
We also notice how both annotators and all models assign very close or the same middle rank scores to three very similar candidates (\textit{Right}, \textit{Uh-huh} and \textit{Oh, sure.}), which indeed all have the same DA (``acknowledgment'').

\end{document}